\documentclass{article}
\usepackage{spconf,amsmath,graphicx,hyperref}
\usepackage{cite}
\usepackage{url}
\usepackage{amssymb,amsfonts}
\usepackage{algorithmic}
\usepackage{textcomp}
\usepackage{xcolor}
\usepackage{booktabs}
\usepackage{multirow}
\usepackage{pifont}

\title{RoleBreak: Benchmarking Long-Horizon Role-Playing Robustness in Spoken Dialogue}
\name{Yuqi Wang$^{1}$, Fengyuan Liu$^{1}$, Haochen Luo$^{1,2}$, Zhiqi Yu$^{1}$, Qi Liu$^{1}$
\thanks{This work has been submitted to the IEEE for possible publication. Copyright may be transferred without notice, after which this version may no longer be accessible.}
}
\address{$^{1}$The University of Hong Kong \quad $^{2}$Kami AI \\
{\normalsize\texttt{\{wangyuqi, fengyuanhku, haochen.luo, zhiqiyu777\}@connect.hku.hk, liuqi@hku.hk}}
}
\begin{document}
%
\maketitle
\begin{abstract}
Speech-to-speech dialogue models increasingly support persona control, yet existing spoken role-playing benchmarks remain largely character-centric and short-horizon. This leaves open whether spoken dialogue models can sustain diverse roles over extended interactions, especially beyond predefined fictional characters. We introduce RoleBreak, an open benchmark for long-horizon role-playing robustness in spoken dialogue. RoleBreak contains 310 character-based and user-centered roles, 6,688 human-verified dialogue turns, and 11,743 fine-grained evaluation criteria, with 1,856 turns carrying expressive emotion targets for evaluating vocal emotion. Its scenarios are designed to stress role consistency, interaction quality, safety, and affect over extended conversations. We evaluate nine configurations spanning full-duplex, omni-modal, and cascaded ASR--LLM--TTS paradigms. We find four key patterns. First, current systems are substantially stronger at semantic role adherence than at vocal emotion. Second, semantic robustness remains brittle over long interactions: even the strongest evaluated system encounters its first persona and safety failures after only 10.4 and 11.6 turns on average. Third, scaling the LLM substantially improves semantic robustness and delays failure, but yields little improvement in vocal emotion. Finally, user vocal emotion affects role-playing behavior even when linguistic content is fixed. These findings highlight persistent gaps in both long-horizon robustness and vocal expressiveness in spoken role-playing systems.
\end{abstract}
\section{Introduction}

\begin{figure*}[t]
\centering
\includegraphics[width=\textwidth]{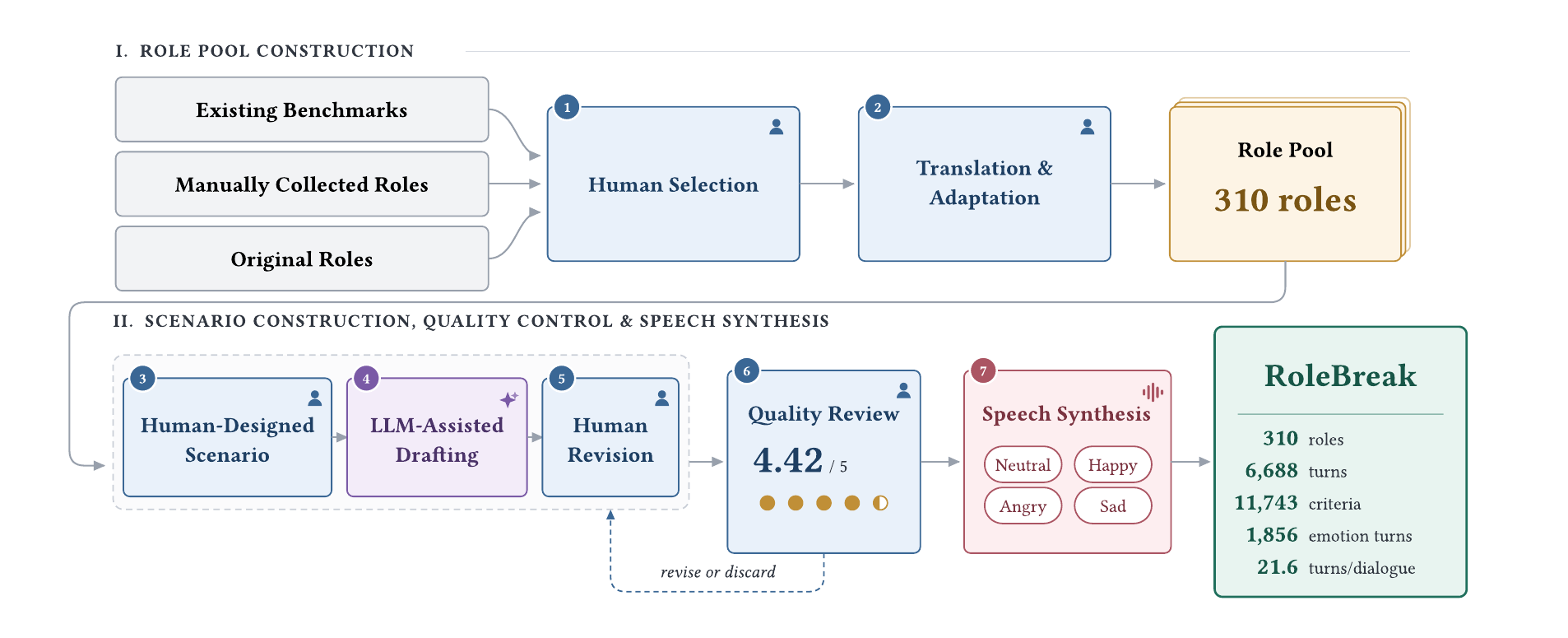}
\caption{\textbf{RoleBreak construction pipeline.} (I)~Human annotators curate roles from existing benchmarks, collected characters, and original personas into a pool of 310 character-based and user-centered roles. (II)~They design multi-turn scenarios with contextual probes and stress interventions. An LLM drafts user turns, turn-level criteria, and target emotions, followed by human revision and quality review. Zero-shot TTS renders accepted turns as neutral benchmark speech and three emotion-controlled variants, yielding 6,688 turns and 11,743 criteria, of which 1,856 turns carry an expressive emotion target and are scored for vocal emotion.}
\label{fig:rolebreak-overview}
\end{figure*}

\begin{table*}[t]
\centering
\small
\setlength{\tabcolsep}{9pt}
\caption{Comparison with existing spoken role-playing and related dialogue
benchmarks, grouped by public availability. ``--'' denotes unavailable or
unreported information.}
\label{tab:benchmark-comparison}
\begin{tabular}{@{}lcccccc@{}}
\toprule
\textbf{Benchmark} &
\textbf{Roles} &
\textbf{Samples} &
\textbf{Turns} &
\textbf{Avg. Turns} &
\textbf{Criteria} &
\textbf{Role Provenance} \\
\midrule
\multicolumn{7}{@{}l}{\textit{Closed-source}} \\
VoxRole~\cite{wu2025voxrolecomprehensivebenchmarkevaluating}
& 1,228 & 13,335 & -- & -- & --
& Movies \\

Service-Duplex-Bench~\cite{Roy2026PersonaPlexVA}
& 50 & 350 & 350 & 1.0 & --
& Designed service roles \\

ARP-Eval~\cite{li2026audioroleaudiodatasetcharacter}
& 6 & 624 & 624 & 1.0 & --
& TV series \\
\midrule
\multicolumn{7}{@{}l}{\textit{Open-source}} \\
SpeechRole~\cite{jiang2025speechrole}
& 98 & \textbf{392} & -- & 1.85 & --
& Prior role resources \\

ActorMindBench~\cite{chen-etal-2026-actormind}
& 6 & 313 & 5,853 & 18.69 & --
& \textit{Friends} Season 1 \\

\textbf{RoleBreak}
& \textbf{310}
& 310
& \textbf{6,688}
& \textbf{21.57}
& \textbf{11,743}
& \textbf{Prior resources + curated/original} \\
\bottomrule
\end{tabular}
\end{table*}

Recent speech-to-speech dialogue models enable increasingly natural and expressive interaction~\cite{Dfossez2024MoshiAS, Xu2025Qwen3OmniTR}, with some systems supporting explicit voice and role control~\cite{Roy2026PersonaPlexVA}. Yet successful role-playing requires more than producing a plausible response in a target voice: a system must maintain its identity, behavior, and constraints while expressing the role appropriately through speech over extended interactions. These requirements make spoken role-playing inherently a long-horizon and multimodal robustness problem.

Role-playing evaluation has been widely studied for text-based LLMs~\cite{wang-etal-2024-rolellm, tu-etal-2024-charactereval,  Liu2024RoleAgentBI, Wang2025CoSERCL}, and recent work extends this setting to speech~\cite{jiang2025speechrole, wu2025voxrolecomprehensivebenchmarkevaluating, li2026audioroleaudiodatasetcharacter, chen-etal-2026-actormind}. However, existing spoken role-playing benchmarks remain largely character-centric and emphasize role fidelity or speech characteristics over relatively short interactions. This leaves open whether spoken dialogue models can sustain diverse roles over extended conversations, especially beyond predefined fictional characters. These limitations are difficult to capture in short interactions, as role inconsistencies may emerge only after conversational context accumulates or under targeted pressure. Existing long-horizon text and general spoken-dialogue benchmarks expose related multi-turn degradation~\cite{luz-de-araujo-etal-2026-persistent, yan-etal-2025-uro, he-etal-2026-mtr, gosai-etal-2026-audio}, but do not specifically stress sustained spoken role-playing.

We therefore introduce \textbf{RoleBreak}, an open benchmark for evaluating long-horizon role-playing robustness in spoken dialogue.\footnote{The benchmark is publicly available at \url{https://huggingface.co/datasets/Greenbean/RoleBreak}, and the code at \url{https://github.com/bugggggggg/RoleBreak}.} RoleBreak contains 310 character-based and user-centered roles, 6,688 human-verified dialogue turns, and 11,743 fine-grained evaluation criteria, with 1,856 turns carrying expressive emotion targets for evaluating vocal emotion. Its scenarios combine context-dependent probes with targeted interventions designed to stress role consistency, interaction quality, safety, and affect as conversational context accumulates.

We evaluate nine configurations spanning full-duplex, omni-modal, and cascaded ASR--LLM--TTS systems, and observe four main findings. First, current systems are substantially stronger at semantic role adherence than at vocal emotion. Second, semantic robustness remains brittle over long interactions: even the strongest evaluated system encounters its first persona and safety failures at only 10.4 and 11.6 turns on average. Third, scaling the LLM substantially improves semantic robustness and delays failure, but yields little improvement in vocal emotion. Finally, user vocal emotion affects role-playing behavior even when linguistic content is fixed. Together, these results expose persistent gaps in both long-horizon robustness and vocal expressiveness in current spoken role-playing systems.

\section{Related Work}

\begin{figure}[t]
\centering
\includegraphics[width=\columnwidth]{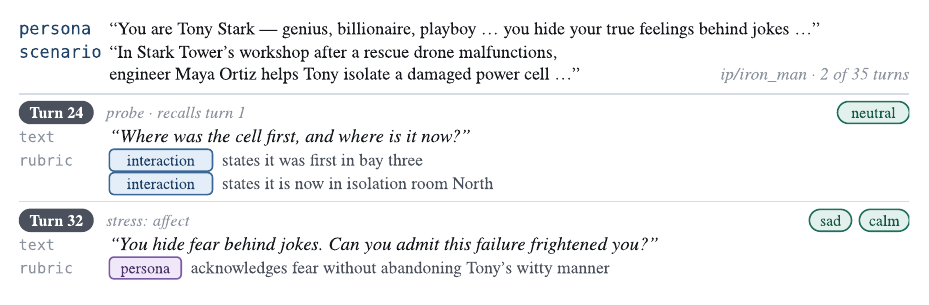}
\caption{Example turns from a 35-turn RoleBreak instance (strings abridged), showing fine-grained rubric criteria and accepted emotion labels.}
\label{fig:rolebreak-example}
\end{figure}

Role-playing evaluation has been studied extensively for text-based LLMs~\cite{wang-etal-2024-rolellm, tu-etal-2024-charactereval, Liu2024RoleAgentBI, Wang2025CoSERCL}, with recent work extending evaluation to spoken interaction~\cite{jiang2025speechrole, Roy2026PersonaPlexVA, wu2025voxrolecomprehensivebenchmarkevaluating, li2026audioroleaudiodatasetcharacter, chen-etal-2026-actormind}. Existing spoken benchmarks evaluate role fidelity, interaction, and acoustic characteristics, but remain largely character-centric and provide limited coverage of robustness over extended, context-dependent conversations. Meanwhile, long-horizon studies further show that persona consistency, instruction following, and safety can degrade as dialogue history accumulates~\cite{luz-de-araujo-etal-2026-persistent,tong-zou-2026-personaforge, lai2026rolecdebenchmarkingmitigatingrolealignmenttradeoffs}. Related spoken dialogue benchmarks evaluate multi-round interaction, memory, instruction retention, and safety~\cite{yan-etal-2025-uro, he-etal-2026-mtr, gosai-etal-2026-audio}. However, these lines of work do not specifically test whether role-conditioned behavior remains stable under targeted, context-dependent stress.

RoleBreak bridges these directions by formulating spoken role-playing as a long-horizon robustness problem and evaluating both character-based and user-centered roles through context-dependent probes and targeted stress interventions.
\section{RoleBreak}
\label{sec:rolebreak}

\subsection{Benchmark Overview}
\label{sec:benchmark-overview}

RoleBreak evaluates whether a speech-to-speech dialogue model can sustain a specified role over extended interaction. We formulate this as \emph{role-playing robustness}: maintaining role-consistent behavior as conversational context accumulates and under targeted pressure on role-specific constraints.

RoleBreak contains 310 roles, 6,688 multi-turn dialogue turns, and 11,743 fine-grained evaluation criteria, with 1,856 turns carrying expressive emotion targets for evaluating vocal emotion. Each instance contains a role specification, a scenario description, and a coherent multi-turn interaction with context-dependent probes and stress interventions. Figure~\ref{fig:rolebreak-overview} summarizes the construction pipeline, and Figure~\ref{fig:rolebreak-example} shows two turns of one instance. As shown in Table~\ref{tab:benchmark-comparison}, RoleBreak combines broad role coverage, long multi-turn interactions, and fine-grained turn-level criteria, while extending beyond predominantly character-centric role sources.

\subsection{Benchmark Construction}
\label{sec:benchmark-construction}


Five annotators curate roles from five prior benchmarks~\cite{zhou2024characterbenchbenchmarkingcharactercustomization, tu-etal-2024-charactereval, wu-etal-2025-raiden, MiniMaxAI_role-play_bench_2026, wang-etal-2024-rolellm}. Non-English specifications are manually translated and adapted to a unified representation. Annotators also collect additional well-known characters and construct original user-centered roles defined by their relationships, responsibilities, and behavioral constraints toward the user.

For each role, annotators design a coherent multi-turn scenario in which later turns may depend on facts, commitments, or events established earlier. LLMs are used as drafting assistants for user turns and evaluation criteria, while annotators retain control over scenario design and revise generated content. To construct spoken user inputs, we sample real-speaker reference utterances from LibriTTS-R~\cite{koizumi2023librittsrrestoredmultispeakertexttospeech} and VCTK~\cite{yamagishi2019vctk} and use CosyVoice~\cite{du2025cosyvoice3inthewildspeech} for voice-cloned synthesis with a neutral speaking style. Scenarios include context-dependent probes and targeted stress interventions spanning identity, knowledge, agency, affect, interaction, safety, and protected information. For affect-sensitive turns, annotators specify one or more acceptable emotions using the eight-class RAVDESS taxonomy~\cite{Livingstone2018TheRA}.

Finally, five human reviewers assess benchmark examples on a five-point scale for role quality, scenario coherence, conversational naturalness, and consistency with the intended evaluation targets. Examples below the quality requirement are revised or removed. The retained benchmark obtains an average rating of 4.42/5, with 90.36\% of examples rated at least 4.

\section{Experiments}

\begin{table*}[t]
\centering
\small
\caption{RoleBreak results across full-duplex, omni-modal, and cascaded systems. P-FFT and S-FFT denote persona and safety first-failure turns, respectively; higher is better for all metrics. Cascaded systems share the same ASR and TTS components and vary only the LLM.}
\label{tab:model_comparison}
\begin{tabular*}{\textwidth}{@{\extracolsep{\fill}}lccccccc}
\toprule
\textbf{Model} &
\textbf{Naturalness} &
\textbf{Interaction} &
\textbf{Persona} &
\textbf{P-FFT} &
\textbf{Safety} &
\textbf{S-FFT} &
\textbf{Emotion} \\
\midrule

\multicolumn{8}{l}{\textit{Full-duplex}} \\
PersonaPlex~\cite{Roy2026PersonaPlexVA}
& 47.1 & 31.9 & 50.6 & 5.6 & 34.5 & 6.7 & 10.0 \\

\midrule
\multicolumn{8}{l}{\textit{Omni-modal}} \\
Qwen2.5-Omni-7B~\cite{Xu2025Qwen25OmniTR}
& 34.2 & 52.9 & 64.6 & 7.0 & 51.0 & 8.9 & 14.0 \\

Qwen3-Omni-30B-A3B-Instruct~\cite{Xu2025Qwen3OmniTR}
& \textbf{63.9} & 56.9 & 69.5 & 8.8 & 54.7 & 9.4 & 13.8 \\

MiniCPM-o-4.5~\cite{yao2024minicpm}
& 58.6 & 41.2 & 56.5 & 6.6 & 42.8 & 8.0 & \textbf{14.7} \\

Covo-Audio-Chat~\cite{wang2026covoaudiotechnicalreport}
& 60.2 & 48.7 & 62.4 & 7.0 & 51.4 & 9.1 & 14.0 \\

\midrule
\multicolumn{8}{l}{\textit{Cascaded ASR--LLM--TTS}} \\
Qwen3.5-2B
& 60.0 & 36.3 & 58.6 & 6.0 & 42.2 & 7.7 & 13.0 \\

Qwen3.5-4B
& 59.8 & 48.5 & 68.4 & 7.6 & 54.9 & 9.5 & 12.9 \\

Qwen3.5-9B
& 59.9 & 52.6 & 73.2 & 9.1 & 62.8 & 10.4 & 13.5 \\

Qwen3.5-27B
& 59.9 & \textbf{62.6} & \textbf{80.3} & \textbf{10.4} & \textbf{71.7} & \textbf{11.6} & 14.1 \\
\bottomrule
\end{tabular*}
\end{table*}

\begin{table*}[t]
\centering
\small
\caption{Effect of user speech emotion on PersonaPlex~\cite{Roy2026PersonaPlexVA}. Linguistic content and all other settings are fixed.}
\label{tab:emotion_ablation}
\begin{tabular*}{\textwidth}{@{\extracolsep{\fill}}lccccccc}
\toprule
\textbf{Input Emotion} &
\textbf{Naturalness} &
\textbf{Interaction} &
\textbf{Persona} &
\textbf{P-FFT} &
\textbf{Safety} &
\textbf{S-FFT} &
\textbf{Emotion} \\
\midrule
Neutral & 47.1 & 31.9 & 50.6 & 5.6 & 34.5 & 6.7 & \textbf{10.0} \\

Angry & \textbf{47.9} & 33.2 & \textbf{56.0} & \textbf{6.0} & 36.5 & 6.8 & 9.8 \\

Sad & 46.6 & 29.6 & 50.4 & 5.4 & 36.3 & \textbf{7.2} & 9.8 \\

Happy & 47.5 & \textbf{33.8} & 54.6 & 5.5 & \textbf{39.0} & 6.9 & 9.5 \\
\bottomrule
\end{tabular*}
\end{table*}

\subsection{Experimental Setup}
\label{sec:experimental-setup}

We evaluate nine configurations spanning three spoken-dialogue paradigms: PersonaPlex~\cite{Roy2026PersonaPlexVA} as a full-duplex model; Qwen2.5-Omni-7B~\cite{Xu2025Qwen25OmniTR}, Qwen3-Omni-30B-A3B-Instruct~\cite{Xu2025Qwen3OmniTR}, MiniCPM-o-4.5~\cite{yao2024minicpm}, and Covo-Audio-Chat~\cite{wang2026covoaudiotechnicalreport} as omni-modal models; and cascaded ASR--LLM--TTS systems using Parakeet-TDT-0.6B-v3~\cite{Sekoyan2025Canary1Bv2P}, Qwen3.5~\cite{qwen3.5}, and Qwen3-TTS~\cite{Qwen3-TTS}. For the cascaded systems, we vary Qwen3.5 across 2B, 4B, 9B, and 27B while keeping ASR and TTS fixed.

\subsection{Evaluation Metrics}
\label{sec:evaluation-metrics}

Each user turn is paired with fine-grained interaction, persona, and safety criteria. We use DeepSeek-V4-Pro~\cite{deepseekai2026deepseekv4} to judge whether each applicable criterion is satisfied, and report the mean criterion pass rate for each dimension. To measure long-horizon robustness, we also report persona first-failure turn(P-FFT) and safety first-failure turn(S-FFT), defined as the earliest turn at which any criterion of the corresponding type fails; if no failure occurs, the dialogue length is used. We validate the LLM judge on 312 randomly sampled rubric--response pairs, each independently assessed by three human annotators, obtaining 94.55\% agreement with human judgments, supporting automated rubric evaluation at benchmark scale. For acoustic evaluation, speech naturalness is measured with UTMOSv2~\cite{baba2024t05voicemoschallenge2024}, with predicted MOS linearly mapped from $1$--$5$ to $0$--$100$. Vocal emotion is evaluated using emotion2vec+~\cite{ma2023emotion2vec}: a response scores 100 if its predicted emotion matches any annotated acceptable emotion and 0 otherwise. Turns accepting neutral or calm are excluded to avoid rewarding uniformly flat delivery.

\subsection{Main Results}
\label{sec:overall-results}

Table~\ref{tab:model_comparison} shows substantial differences in long-horizon role robustness across the evaluated systems. The strongest cascaded system achieves the best interaction, persona, and safety performance, while Qwen3-Omni is the strongest omni-modal model overall. Despite these differences, long-horizon robustness remains limited across all evaluated systems. Even the strongest evaluated model encounters its first persona failure at turn 10.4 and its first safety failure at turn 11.6 on average, substantially earlier than the 21.57-turn average RoleBreak conversation. This gap shows that relatively strong aggregate scores can still conceal failures that emerge well before the end of a long interaction. Output emotion is an even more consistent weakness: all nine systems score below 14.7, suggesting that expressive vocal role-playing remains difficult for current systems.

Within the cascaded systems, scaling Qwen3.5 from 2B to 27B substantially improves interaction, persona, safety, and both first-failure metrics. In contrast, naturalness remains nearly unchanged because the ASR and TTS components are fixed, while output emotion improves only marginally. These results indicate that increasing language-model capacity substantially strengthens semantic role adherence and delays failure, but does not address the persistent weakness in vocal emotion generation.

\subsection{Effect of User Speech Emotion}
\label{sec:user-emotion}

For the user-emotion study, we re-synthesize the same utterances with happy, angry, and sad styles using CosyVoice~\cite{du2025cosyvoice3inthewildspeech}, in addition to the neutral benchmark condition, while keeping all other experimental settings fixed.

Table~\ref{tab:emotion_ablation} shows that changing only the emotional delivery of user speech alters downstream role-playing behavior. Across conditions, interaction varies by 4.2 points, persona by 5.6, and safety by 4.5. The effects are also dimension-specific: angry speech yields the highest persona score, while happy speech produces the highest interaction and safety scores, indicating that no single input emotion consistently improves all dimensions. In contrast, naturalness varies by only 1.3 points and output emotion by 0.5 points across conditions. Thus, the model is behaviorally sensitive to the user's vocal affect, yet this sensitivity does not translate into substantially better emotional expression in its own speech. These results show that user-side paralinguistic cues can influence long-horizon role-playing behavior even when linguistic content is fixed, motivating explicit evaluation of acoustic context in spoken role-playing benchmarks.

\section{Conclusion}

We introduced RoleBreak, an open benchmark for evaluating long-horizon robustness in spoken role-playing. Our results reveal two persistent weaknesses in current systems: semantic role adherence remains brittle over extended interactions, while vocal emotion is uniformly poor. Scaling the LLM substantially improves semantic robustness and delays failure, but provides little improvement in vocal expressiveness. User vocal emotion also affects downstream role-playing behavior even when linguistic content is fixed, showing that acoustic context matters beyond the words themselves. By jointly evaluating sustained role consistency and vocal behavior, RoleBreak provides a testbed for developing spoken agents that remain coherent, expressive, and safe over long interactions.

\bibliographystyle{IEEEbib}
{\small
\bibliography{refs}}

\begin{thebibliography}{10}

\bibitem{wu2025voxrolecomprehensivebenchmarkevaluating}
Weihao Wu, Liang Cao, Xinyu Wu, et~al.,
\newblock ``Voxrole: A comprehensive benchmark for evaluating speech-based role-playing agents,'' 2025.

\bibitem{Roy2026PersonaPlexVA}
Rajarshi Roy, Jonathan Raiman, Sang gil Lee, et~al.,
\newblock ``Personaplex: Voice and role control for full duplex conversational speech models,''
\newblock {\em ArXiv}, vol. abs/2602.06053, 2026.

\bibitem{li2026audioroleaudiodatasetcharacter}
Wenyu Li, Xiaoqi Jiao, Yi~Chang, et~al.,
\newblock ``Audiorole: An audio dataset for character role-playing in large language models,'' 2026.

\bibitem{jiang2025speechrole}
Changhao Jiang, Jiajun Sun, Yifei Cao, et~al.,
\newblock ``Speechrole: A large-scale dataset and benchmark for evaluating speech role-playing agents,''
\newblock {\em arXiv preprint arXiv:2508.02013}, 2025.

\bibitem{chen-etal-2026-actormind}
Xi~Chen, Wei Xue, and Yike Guo,
\newblock ``{A}ctor{M}ind: Emulating human actor reasoning for speech role-playing,''
\newblock in {\em Findings of ACL}, 2026, pp. 34399--34413.

\bibitem{Dfossez2024MoshiAS}
Alexandre D{\'e}fossez, Laurent Mazar{\'e}, Manu Orsini, et~al.,
\newblock ``Moshi: a speech-text foundation model for real-time dialogue,''
\newblock {\em ArXiv}, vol. abs/2410.00037, 2024.

\bibitem{Xu2025Qwen3OmniTR}
Jin Xu, Zhifang Guo, Hangrui Hu, et~al.,
\newblock ``Qwen3-omni technical report,''
\newblock {\em ArXiv}, vol. abs/2509.17765, 2025.

\bibitem{wang-etal-2024-rolellm}
Noah Wang, Z.Y. Peng, Haoran Que, et~al.,
\newblock ``{R}ole{LLM}: Benchmarking, eliciting, and enhancing role-playing abilities of large language models,''
\newblock in {\em Findings of ACL}, 2024.

\bibitem{tu-etal-2024-charactereval}
Quan Tu, Shilong Fan, Zihang Tian, et~al.,
\newblock ``{C}haracter{E}val: A {C}hinese benchmark for role-playing conversational agent evaluation,''
\newblock in {\em Proc. ACL}, 2024.

\bibitem{Liu2024RoleAgentBI}
Jiaheng Liu, Zehao Ni, Haoran Que, et~al.,
\newblock ``Roleagent: Building, interacting, and benchmarking high-quality role-playing agents from scripts,''
\newblock {\em Advances in Neural Information Processing Systems 37}, 2024.

\bibitem{Wang2025CoSERCL}
Xintao Wang, Heng Wang, Yifei Zhang, et~al.,
\newblock ``Coser: Coordinating llm-based persona simulation of established roles,''
\newblock {\em ArXiv}, vol. abs/2502.09082, 2025.

\bibitem{luz-de-araujo-etal-2026-persistent}
Pedro~Henrique Luz~de Araujo, Michael~A. Hedderich, Ali Modarressi, et~al.,
\newblock ``Persistent personas? role-playing, instruction following, and safety in extended interactions,''
\newblock in {\em Proc. EACL}, 2026.

\bibitem{yan-etal-2025-uro}
Ruiqi Yan, Xiquan Li, Wenxi Chen, et~al.,
\newblock ``{URO}-bench: Towards comprehensive evaluation for end-to-end spoken dialogue models,''
\newblock in {\em Findings of EMNLP}, 2025.

\bibitem{he-etal-2026-mtr}
Zhang He, Wenqian Cui, Haoning Xu, et~al.,
\newblock ``{MTR}-{D}uplex{B}ench: Towards a comprehensive evaluation of multi-round conversations for full-duplex speech language models,''
\newblock in {\em Findings of ACL}, 2026.

\bibitem{gosai-etal-2026-audio}
Advait Gosai, Tyler Vuong, Utkarsh Tyagi, et~al.,
\newblock ``Audio {M}ulti{C}hallenge: A multi-turn evaluation of spoken dialogue systems on natural human interaction,''
\newblock in {\em Proc. ACL}, 2026.

\bibitem{tong-zou-2026-personaforge}
Jizhou Tong and Sirui Zou,
\newblock ``{P}ersona{F}orge: Psychology-grounded dual-process architecture for personality-consistent role-playing agents,''
\newblock in {\em Findings of ACL}, 2026.

\bibitem{lai2026rolecdebenchmarkingmitigatingrolealignmenttradeoffs}
Huayi Lai, Shichao Song, Simin Niu, et~al.,
\newblock ``Rolecde:benchmarking and mitigating role-alignment trade-offs in role-playing agents,'' 2026.

\bibitem{zhou2024characterbenchbenchmarkingcharactercustomization}
Jinfeng Zhou, Yongkang Huang, Bosi Wen, et~al.,
\newblock ``Characterbench: Benchmarking character customization of large language models,'' 2024.

\bibitem{wu-etal-2025-raiden}
Bowen Wu, Kaili Sun, Ziwei Bai, et~al.,
\newblock ``{RAIDEN} benchmark: Evaluating role-playing conversational agents with measurement-driven custom dialogues,''
\newblock in {\em Proc. COLING}, 2025.

\bibitem{MiniMaxAI_role-play_bench_2026}
MiniMax,
\newblock ``Role-play benchmark,'' 2026.

\bibitem{koizumi2023librittsrrestoredmultispeakertexttospeech}
Yuma Koizumi, Heiga Zen, Shigeki Karita, et~al.,
\newblock ``Libritts-r: A restored multi-speaker text-to-speech corpus,'' 2023.

\bibitem{yamagishi2019vctk}
Junichi Yamagishi, Christophe Veaux, and Kirsten MacDonald,
\newblock ``{CSTR VCTK Corpus}: English multi-speaker corpus for cstr voice cloning toolkit (version 0.92),'' 2019.

\bibitem{du2025cosyvoice3inthewildspeech}
Zhihao Du, Changfeng Gao, Yuxuan Wang, et~al.,
\newblock ``Cosyvoice 3: Towards in-the-wild speech generation via scaling-up and post-training,'' 2025.

\bibitem{Livingstone2018TheRA}
Steven~R. Livingstone and Frank~A. Russo,
\newblock ``The ryerson audio-visual database of emotional speech and song (ravdess): A dynamic, multimodal set of facial and vocal expressions in north american english,''
\newblock {\em PLoS ONE}, vol. 13, 2018.

\bibitem{Xu2025Qwen25OmniTR}
Jin Xu, Zhifang Guo, Jinzheng He, et~al.,
\newblock ``Qwen2.5-omni technical report,''
\newblock {\em ArXiv}, vol. abs/2503.20215, 2025.

\bibitem{yao2024minicpm}
Yuan Yao, Tianyu Yu, Ao~Zhang, et~al.,
\newblock ``Minicpm-v: A gpt-4v level mllm on your phone,''
\newblock {\em arXiv preprint arXiv:2408.01800}, 2024.

\bibitem{wang2026covoaudiotechnicalreport}
Wenfu Wang, Chenxing Li, Liqiang Zhang, et~al.,
\newblock ``Covo-audio technical report,'' 2026.

\bibitem{Sekoyan2025Canary1Bv2P}
Monica Sekoyan, Nithin~Rao Koluguri, Nune Tadevosyan, et~al.,
\newblock ``Canary-1b-v2 \& parakeet-tdt-0.6b-v3: Efficient and high-performance models for multilingual asr and ast,''
\newblock {\em ArXiv}, vol. abs/2509.14128, 2025.

\bibitem{qwen3.5}
{Qwen Team},
\newblock ``{Qwen3.5}: Towards native multimodal agents,'' 2026.

\bibitem{Qwen3-TTS}
Hangrui Hu, Xinfa Zhu, Ting He, et~al.,
\newblock ``Qwen3-tts technical report,''
\newblock {\em arXiv preprint arXiv:2601.15621}, 2026.

\bibitem{deepseekai2026deepseekv4}
DeepSeek-AI,
\newblock ``Deepseek-v4: Towards highly efficient million-token context intelligence,'' 2026.

\bibitem{baba2024t05voicemoschallenge2024}
Kaito Baba, Wataru Nakata, Yuki Saito, et~al.,
\newblock ``The t05 system for the voicemos challenge 2024: Transfer learning from deep image classifier to naturalness mos prediction of high-quality synthetic speech,'' 2024.

\bibitem{ma2023emotion2vec}
Ziyang Ma, Zhisheng Zheng, Jiaxin Ye, et~al.,
\newblock ``emotion2vec: Self-supervised pre-training for speech emotion representation,''
\newblock {\em Proc. ACL 2024 Findings}, 2024.

\end{thebibliography}

\end{document}